%% file: main.tex
\pdfoutput=1

\documentclass[11pt]{article}

\usepackage{EMNLP2022} 

\usepackage{times}
\usepackage{latexsym}

\usepackage[T1]{fontenc}

\usepackage[utf8]{inputenc}

\usepackage{microtype}

\usepackage{inconsolata}

\usepackage{csquotes}
\usepackage{array} 
\usepackage{graphicx}
\usepackage{float}
\usepackage{comment}
\usepackage{caption}
\usepackage{multirow}

\title{Few-Shot Out-of-Domain Transfer Learning of Natural Language Explanations in a Label-Abundant Setup}

\author{
  \textbf{Yordan Yordanov}\textsuperscript{1}, 
   \textbf{Vid Kocijan}\textsuperscript{2},  
  \textbf{Thomas Lukasiewicz}\textsuperscript{3,1}, 
  \textbf{Oana-Maria Camburu}\textsuperscript{4} \\
  \textsuperscript{1}\,University of Oxford 
  \hspace{1pt}
  \textsuperscript{2}\,Kumo.ai 
  \hspace{1pt}
  \textsuperscript{3}\,TU Wien \\
  \textsuperscript{4}\,University College London \\
  \texttt{yordan.yordanov@cs.ox.ac.uk}, \hspace{5pt}\texttt{thomas.lukasiewicz@tuwien.ac.at}, \\
  \texttt{vid@kumo.ai}, \hspace{3pt}
  \texttt{o.camburu@cs.ucl.ac.uk}
\\
}
\begin{document}
\maketitle

\begin{abstract}

Training a model to provide natural language explanations (NLEs) for its predictions usually requires the acquisition of task-specific NLEs, which is time- and resource-consuming. A potential solution is the few-shot out-of-domain transfer of NLEs from a parent task with many NLEs to a child task.
In this work, we examine the setup in which the child task has few NLEs but abundant labels. We establish four few-shot transfer learning methods that cover the possible fine-tuning combinations of the labels and NLEs for the parent and child tasks.
We transfer explainability from a large natural language inference dataset (e-SNLI) separately to two child tasks: (1) hard cases of pronoun resolution, where we introduce the small-e-WinoGrande dataset of NLEs on top of the WinoGrande dataset, and (2)~commonsense validation (ComVE). Our results demonstrate that the parent task helps with NLE generation and we establish the best methods for this setup.

\end{abstract}

\section{Introduction}

Recent developments have made it possible for AI models to learn from natural language explanations (NLEs) for the ground-truth labels at training time and generate such explanations for their decisions at deployment time \citep{DBLP:conf/eccv/HendricksARDSD16, math, zeynep, esnli, cars, rajani-etal-2019-explain_2, inconsistencies, narang2020wt5, kumar-talukdar-2020-nile_2, marasovi2022fewshot}. 
However, large datasets of NLEs, such as e-SNLI \citep{esnli}, are time-consuming and expensive to gather. One approach is to transfer explanations from a different domain, via few-shot transfer learning. The usual setup for few-shot out-of-domain transfer learning consists of transfer learning from a \enquote{parent} task, with abundant training examples, to a \enquote{child} task that only has a few training examples \citep{few-shot-old, Ravi2017OptimizationAA}. 

In this work, we assume that the child task has few training NLEs but abundant labels. 
Given the advent of deep learning in the last years, this scenario may be quite frequent, as one may already have a large dataset with labels on which they aim to train NLEs-generating models without annotating the entire dataset with NLEs. 
To our knowledge, there is only one existing work in this setup, that of \citet{cross-domain_transfer_nles}, which introduces a vanilla fine-tuning method on top of the zero-shot WT5 model \citep{narang2020wt5}. 
However, their work is limited by: (1)  they only test one of the four possible scenarios we identify for this setup, and (2) they use only automatic evaluation metrics, which do not necessarily align with human judgment \citep{esnli, kayser2021evil}.  

In this work, we introduce three few-shot transfer learning methods for NLEs that utilize the abundant training labels for both the parent and child task, and we adapt for computational efficiency the method from \citet{cross-domain_transfer_nles}. Together, these four methods are combinations of multi-task learning and fine-tuning between a parent and a child task with few training NLEs but abundant labels. We instantiate our few-shot learning approaches on e-SNLI \citep{esnli} as parent task, and WinoGrande \citep{winogrande} and ComVE \citep{wang-etal-2020-semeval_2} as child tasks. As the WinoGrande dataset does not contain NLEs, we introduce small-e-WinoGrande, which provides 100/50/100 NLEs for the training, development, and test sets, respectively. We show the extent to which few-shot out-of-domain transfer learning of NLEs is currently feasible, and provide insight into which learning techniques work best in this setup. We perform human evaluation and compare against child-task-only and zero-shot baselines.
\footnote{The code and the datasets are publicly available at: \url{https://github.com/YDYordanov/Few-shot-NLEs}.}

\input{Experimental_Setup}

\input{Results}

\section{Related Work}

There are three main focuses in NLE generation: quality improvement \citep{esnli, narang2020wt5, valentino-etal-2022-case}, NLE faithfulness \citep{kumar-talukdar-2020-nile_2, wiegreffe-etal-2021-measuring_2, liu-etal-2019-towards-explainable_2, latcinnik2020explaining}, and transfer learning of NLEs. Zero-shot in-domain transfer of NLEs (between datasets of the same task) has been done, e.g., by~\citet{esnli, kumar-talukdar-2020-nile_2}, and~\citet{narang2020wt5}. \citet{narang2020wt5} additionally consider zero-shot out-of-domain transfer of NLEs, while \citet{cross-domain_transfer_nles} extend their work by showing that few-shot out-of-domain transfer of NLEs is possible in the abundant-label setup. \citet{marasovi2022fewshot} use prompt engineering for few-shot out-of-domain transfer of NLEs for the scarce-label setup. The prompt choice is less relevant in our abundant-label setup, because task adaptation can be done via the abundant training labels. 
In the more general area of natural language generation, few-shot learning is a growing topic \citep{chen-etal-2020-shot_2}, e.g., in dialog generation \citep{peng-etal-2020-shot_2, shalyminov-etal-2019-shot_2}. These approaches, however, do not directly apply to transfer learning of NLEs, which is a dual task of predicting both the label and generating an explanation. 




\section{Summary and Outlook}

In this work, we investigated four methods for few-shot out-of-domain transfer learning of NLEs for the abundant-label setting. 
We introduced small-e-WinoGrande, a dataset of NLEs on top of a small sample of instances from WinoGrande.
We showed that out-of-domain few-shot learning can significantly help with NLE generation compared to zero-shot or child-task-only learning. Amongst the four NLE few-shot learning methods, we found that the most convincing NLEs are generated by the methods that provide separate training regimes for the child task and its few training NLEs. 
While our results indicate that few-shot out-of-domain transfer learning of NLEs is helpful, there is room for improvement both in the quality of the generated NLEs and in task-performance. Thus, our work 
provides an essential foundation for future research into few-shot out-of-domain transfer learning of NLEs where label abundance is available. 

\section{Limitations}

The training methods in this work can apply to any language other than English, but a large parent task with NLEs is needed and a high-performance pre-trained generative language model may be needed for that language. Training one of our methods takes approximately three hours per e-SNLI epoch on one NVIDIA TITAN Xp GPU, which should be multiplied by the number of epochs and the hyperparameter combinations used. In practice, we observed that the results are sensitive to the number of epochs and to the choice of the learning rate, so a comprehensive hyperparameter search may be needed. This significantly increases the computational requirements and can be an obstacle for researchers on a limited budget. In total, the required time to reproduce all our results is approximately 45 GPU days. Scaling our methods to larger language models can also be challenging from a computational requirements standpoint.

\section{Acknowledgments}
This work was supported by an Early Career Leverhulme Fellowship, by the Alan Turing Institute under the EPSRC grant EP/N510129/1, by the AXA
Research Fund, and by the EPSRC Studentship OUCS/EPSRC-NPIF/VK/1123106.
We also acknowledge the use of the EPSRC-funded Tier 2
facility JADE (EP/P020275/1) and GPU computing support
by Scan Computers International Ltd.
\bibliography{references}
\bibliographystyle{acl_natbib}

\input{Appendices}

\end{document}

%% file: Experimental_Setup.tex
\section{Experimental Setup}

\subsection{Datasets}

\paragraph{e-SNLI.} 
Natural language inference (NLI) \citep{rte-1} is the task of assigning a relation of \textit{entailment}, \textit{contradiction}, or \textit{neutrality} between a \textit{premise} and a \textit{hypothesis}. The e-SNLI dataset \citep{esnli} consists of human-written NLEs on top of the Stanford Natural Language Inference (SNLI) 
\citep{bowman-etal-2015-large_2}. 
We select e-SNLI 
as parent dataset due to its large size ($\sim$570K) and high-quality NLEs.

\paragraph{WinoGrande.}
The WinoGrande dataset \citep{winogrande} consists of 40,398 binary fill-in-the-gap instances of pronoun resolution that follow the Winograd Schema format \citep{levesque2012winograd}. We select WinoGrande as a child task, since it requires implicit knowledge, which we want to capture in the NLEs. We construct the \textit{small-e-WinoGrande} dataset by manually creating NLEs for 100/50/100 training/dev/test instances.

\paragraph{ComVE.} 

Commonsense Validation and Explanation (ComVE) \citep{wang-etal-2020-semeval_2}, as reformulated by \citet{majumder2021rationaleinspired}, is the task of jointly identifying which one of two statements contradicts commonsense and explaining why. The dataset consists of 10,000 training, 1,000 validation, and 1,000 test instances. 
We select ComVE as a child task, because it is a commonsense reasoning task for which there are good-quality human-written NLEs. 
%
For more dataset details, see Appendix~\ref{datasets}.

\subsection{Base Model}
\label{model_section}

\begin{table}
    \centering
    \scalebox{0.8}{
    \begin{tabular}{l
    >{\arraybackslash}m{4.0cm}
    >{\arraybackslash}m{2.9cm}}
    
    \hline
    \textbf{Task} & \textbf{Input Format} & \textbf{Target Format} \\
    \hline
    e-SNLI & explain nli premise: [premise] hypothesis: [hypothesis] & [relation] explanation: [explanation] \\
    \hline
    W.G. & explain WinoGrande schema: [schema start] \underline{\hspace{0.3cm}} [schema end] options: [option 1], [option 2]. & [correct option] explanation: [explanation] \\
    \hline
    ComVE & explain ComVE Sentence 1: [statement 1] Sentence 2: [statement 2] & [nonsensical statement id] explanation: [explanation] \\
    \hline

    \end{tabular}}
    \caption{T5 input/target formats for each task, used for all models. When training on examples without NLEs, \enquote{explain} and \enquote{explanation:} are not included in the input/target format.}
    \label{tab:input_formats}
\end{table}

Similarly to \citet{narang2020wt5}, we use the T5 \citep{t5} generative language model, in particular, the \enquote{Base} model with 220M parameters, due to its good trade-off of performance and computational requirements. For T5, tasks are distinguished only via their task-specific input/target formats. We follow the input/target format for e-SNLI by \citet{narang2020wt5}: \textit{premise: [premise] hypothesis: [hypothesis] / [relation] explanation: [explanation]}. We obtain the input formats for WinoGrande and ComVE in a similar manner (see Table~\ref{tab:input_formats}). We observed in early experiments that the exact choice of input/target formats does not significantly affect performance. 

We choose the best practice for multi-task learning with T5, namely, via training on the union of the datasets in question \citep{t5}.

\subsection{Few-Shot Transfer Learning Methods}

\begin{table*}
    \centering
    \scalebox{0.85}{
    \begin{tabular}{ll}
    \hline
    \textbf{Model name} & \textbf{Meaning} \\
    \hline
    CD--fine-tune & fine-tune T5 on the child dataset, and then fine-tune on 50 NLEs \\
    CD--union & fine-tune T5 on the union of the child dataset and 50 NLEs \\
    WT5--fine-tune & fine-tune T5 on the union of e-SNLI and SNLI, and then fine-tune on the child dataset \\
    WT5 & fine-tune T5 on the union of e-SNLI, SNLI, and the child dataset \\
    
    \hline
    
    M1 & fine-tune T5 on the union of e-SNLI, the child dataset, and 50 NLEs \\
    M2 & fine-tune T5 on the union of e-SNLI and the child dataset, and then fine-tune on 50 NLEs \\
    M3 & fine-tune T5 on e-SNLI, and then fine-tune on the union of the child dataset and 50 NLEs \\
    M4 & fine-tune T5 on e-SNLI, and then fine-tune on the child dataset, and, finally, on 50 NLEs \\
    
    \hline
    \end{tabular}}
    \caption{Legend of the model names. The child dataset excludes the NLEs, unless specified. The 50 NLEs refer to the few (50) instances of the child task with NLEs.}
    \label{tab:model_legend}
\end{table*}

Table~\ref{tab:model_legend} shows all the models that we use. M1 to M4 are the four few-shot transfer learning methods for NLE generation, which we obtain by combining the parent dataset with NLEs, the child dataset, and a few NLEs (we use 50 in this work) in all reasonable multi-task and fine-tuning combinations. M3 is similar to the method by \citet{cross-domain_transfer_nles}, but the latter uses the union of the parent dataset with and without explanations, mimicking WT5. We choose against this, because in the few-shot NLE case, this is unnecessary and doubles the computation cost. 

We also consider four baseline methods. 
The two child-task baselines CD--fine-tune and CD-union serve to measure the contribution of the parent in NLE transfer. Two zero-shot NLE transfer learning baselines, WT5 \citep{narang2020wt5} and WT5--fine-tune, serve to measure the contribution of the 50 training NLEs in the child task. 
The training details are given in Appendix~\ref{hyperparams}.

\subsection{Human Evaluation}

\begin{table*}
\centering
\scalebox{0.9}{
\begin{tabular}{m{2.4cm}
>{\centering\arraybackslash}m{0.9cm}
>{\centering\arraybackslash}m{0.9cm}
>{\centering\arraybackslash}m{0.9cm}
>{\centering\arraybackslash}m{0.9cm}
>{\centering\arraybackslash}m{0.8cm}
>{\centering\arraybackslash}m{0.8cm}
>{\centering\arraybackslash}m{0.8cm}
>{\centering\arraybackslash}m{0.8cm}
>{\centering\arraybackslash}m{1.5cm}
>{\centering\arraybackslash}m{1.9cm}}
\hline
    \multirow{2}{*}{\textbf{Model}} & \multicolumn{2}{c}{\textbf{WinoGrande}} & \multicolumn{2}{c}{\textbf{ComVE}} & \multicolumn{6}{c}{\textbf{ComVE Automatic NLE Metrics}} \\
    & \textbf{Task acc\%} & \textbf{NLE score} & \textbf{Task acc\%} & \textbf{NLE score} & \textbf{B-1} & \textbf{B-2} & \textbf{B-3} & \textbf{B-4} & \textbf{METEOR} & \textbf{BERTScore} \\
    \hline
    CD--fine-tune & 59.7 & 34.7 & \textbf{87.8} & 31.4 & \textbf{45.2} & \textbf{29.5} & \textbf{19.5} & \textbf{13.1} & \textbf{21.5} & 83.4 \\
    CD--union & 57.2 & 35.9 & 83.1 & 27.7 & 27.4 & 16.6 & 10.2 & 6.4 & 19.1 & 81.8 \\
    WT5--fine-tune & \textbf{60.2} & 8.7 & 85.7 & 28.9 & 24.6 & 15.1 & 9.7 & 6.5 & 13.5 & 74.8 \\
    WT5 & 58.0 & 8.3 & 76.2 & 23.9 & 22.8 & 12.0 & 6.4 & 3.6 & 12.7 & 71.5 \\
    \hline
    M1 & 53.6 & 28.3 & 82.8 & 40.2 & 34.5 & 19.2 & 10.8 & 6.3 & 20.3 & 81.8 \\
    M2 & 56.0 & \textbf{44.1}* & 80.6 & 40.6 & 43.5 & 26.3 & 16.5 & 10.6 & 20.0 & 83.1 \\
    M3 & 54.6 & 29.6 & 85.5 & 38.6 & 33.6 & 18.8 & 10.9 & 6.2 & 20.8 & 82.1 \\
    M4 & 58.2 & 41.9* & 86.5 & \textbf{48.5}* & 44.4 & 27.5 & 17.5 & 10.7 & 21.2 & \textbf{83.6} \\
    
\hline
\end{tabular}}
\caption{\label{tab:wg_comve_models}
Performance of models on WinoGrande and ComVE as child tasks. From the 100 test examples, only the correctly classified are given NLE scores. B-1,2,3,4 stand for BLEU-1,2,3,4. Best results are in bold; * denotes the statistically significant best results. 
}
\end{table*}

We use Amazon Mechanical Turk to evaluate the model-generated NLEs, with three annotators per instance. The evaluation procedure for each instance is in three steps and follows existing works \citep{kayser2021evil, majumder2021rationaleinspired, marasovi2022fewshot}. First, annotators have to predict the classification label for the example. Second, they have to select one of four options for whether the NLE is a valid and satisfactory explanation for the selected label: Yes, Weak~Yes, Weak~No, or No. Third, they have to select shortcomings of the explanation from the following: \enquote{does not make sense}, \enquote{insufficient justification}, \enquote{irrelevant to the task}, \enquote{too trivial}, and \enquote{none}. 



All models are evaluated on 100 examples from the test dataset of each child task. 
Similarly to previous works~\citep{esnli, kayser2021evil, majumder2021rationaleinspired}, the NLE evaluation is only done on correctly labeled (by the model) examples, as it is expected that an incorrect label is not supported by the model with a correct NLE. See Appendix~\ref{appendix:human_evaluation} for more details and for screenshots of the forms used to collect the annotations. 


%% file: Results.tex
\section{Results}



Following \citet{kayser2021evil}, we use an aggregated score (we call \enquote{NLE score}) of the four categories (Yes, Weak~Yes, No, Weak~No) to compare the NLE generation quality, where Yes, Weak~Yes, Weak~No, and No are given the weights $1$, $2/3$, $1/3$, and $0$, respectively. This aggregation has two goals: (1) to provide a single metric to compare the methods, and (2) to account for the subjective nature of choosing between close labels such as Yes and Weak~Yes. A summary of the Yes, Weak~Yes, Weak~No, and No scores and the shortcomings are presented in Appendix~\ref{additional_results_appendix}.

For every model comparison, we report if it is statistically significant via the paired Student's t-test for equal variances \citep{t-test_equal_variances}, with single-tailed p-values and 0.05 statistical significance threshold. 

The results are given in Table~\ref{tab:wg_comve_models}. We only report automatic metrics (BLEU~\citep{papineni-etal-2002-bleu_2}, METEOR~\citep{banerjee-lavie-2005-meteor_2}, and BERTScore~\citep{Zhang*2020BERTScore}) for NLE quality for ComVE, since WinoGrande has only a small number of test NLE instances (which have been used for grounding in our human evaluation -- see Appendix \ref{appendix:human_evaluation}). We notice that the automatic metrics are not well aligned with the human evaluation (NLE score). This has also been previously observed in other studies \cite{esnli, kayser2021evil}. Therefore, we will base our conclusions only on the human evaluation (NLE score). 

First, we notice that all methods (M1--M4) significantly outperform the zero-shot baselines (WT5--fine-tune and WT5) in terms of NLE quality for both datasets, which proves the utility of the 50 child-task NLEs. 

Second, we see that not all methods outperform the child-task baselines. For example, on WinoGrande, both CD--fine-tune and CD-union outperform M1 and M3 in terms of the NLE quality. This shows that it is sometimes possible that fine-tuning on a large parent task of out-of-domain NLEs hurts NLE quality of a child task. However, for both datasets, the best performing method is among the M1--M4 methods (and for ComVE, all M1--M4 methods outperform the child-task baselines), suggesting that it is generally useful to use a large dataset of NLEs as a parent task even when out-of-domain. 

Third, we see that the M1--M4 methods rank differently on different datasets, in particular, M2 and M4 are the significantly best methods on WinoGrande, and M4 is the significantly best method on ComVE. We believe that the main difference in method ranking is that the methods obtain much closer-to-chance accuracy on WinoGrande than on ComVE. This is important, because poor task understanding can lead to poor NLE generation. In particular, M2 and M4 obtain significantly better WinoGrande accuracy than M1 and M3, which might explain the significant gap in NLE score. M2 and M4 are the best NLE generation methods on both datasets, which could be because M2 and M4 use the 50 NLEs of the child task in a separate training regime, whereas M1 and M3 use a combination of the much larger child dataset with the 50 NLEs. This suggests that the 50 NLEs require their own training regime to obtain good NLE generation.

Finally, we see that the best task performances on both datasets are obtained by a baseline method: WT5–fine-tune (no child NLEs) for WinoGrande and CD--fine-tune (child-task only) for ComVE. Moreover, among the M1-M4 methods, there is no clear best method in terms of task performance, nor a correlation with the NLE quality. Hence, a trade-off between task performance and NLE quality is needed, which we observe is achieved by M4. 
\subsection{Qualitative Examples}

\paragraph{WinoGrande.}
While many of the correct NLEs that are generated by the models are trivial, e.g.:

\begin{quote}
    Fiction books were interesting and easy to read for Logan but not Brett
because \underline{\hspace{0.5cm}} enjoyed real tales.
\textbf{A:} Brett. \\
    \textbf{NLE:} Fiction books were interesting and easy to read for
Logan but not Brett because Brett enjoyed real tales
\end{quote}

there are some non-trivial generations, e.g.: 

\begin{quote}
    Brett was browsing the internet while he found the information unlike Randy, \underline{\hspace{0.5cm}} prefers using books. \textbf{A:} Randy. \\
    \textbf{NLE:} If Brett is browsing the internet, then he is more likely to find information than Randy, who prefers to use books.
\end{quote}

\paragraph{ComVE.}
By manual inspection, the correct model-generated NLEs are either a repetition of the correct statement, or a negation of the nonsensical statement. Examples from each: 

\begin{quote}
    \textbf{Statements:} The fire will [burn / cool] you if you touch it. \textbf{A:} burn. \textbf{NLE:} The fire will burn you if you touch it.

    \textbf{Statements:} He inserts [his thumb / the keys] into the car ignition switch. \\ \textbf{A:} the keys. \textbf{NLE:} One cannot insert his thumb into the car ignition switch.
\end{quote}

For more qualitative results, see Appendix~\ref{all_nle_examples}, where six non-cherry-picked NLEs from each model are presented.

%% file: Appendices.tex
\clearpage
\appendix

\section{Datasets}
\label{datasets}

\paragraph{WinoGrande.}
Because of the lack of a publicly available test set (testing happens through its leaderboard,\footnote{\url{https://leaderboard.allenai.org/winogrande/submissions/public}} which has submission limitations), we do a random split of the original WinoGrande training dataset into 39,130 training instances (called WG-train) and 1,268 validation instances (called WG-dev). For testing, we use the original WinoGrande development set, which we denote by WG-test.

We created the small-e-WinoGrande dataset by manually constructing NLEs for 100 examples from WG-train, 50 examples from WG-dev, and 100 examples from WG-test. Example:
\begin{quote}
    The geese prefer to nest in the fields rather than the forests because in the \underline{\hspace{0.5cm}} predators are very visible. \\  \textbf{Options:} fields, forests. \textbf{Answer:} fields. \\ \textbf{NLE:} The fields are more open spaces than the forests, hence predators are more visible there.
\end{quote}

\paragraph{ComVE.} 
Originally, ComVE \citep{wang-etal-2020-semeval_2} consists of three tasks: A, B, and C, where only tasks A and C are relevant for this work. ComVE-A is the classification task of identifying which statement out of a pair of statements does not make sense. The ComVE-C task provides only the statement that does not make sense (from the pair) and requires the model to generate an NLE for why that is the case. 
To form a classification task with explanations, we merge tasks A and C by matching the nonsensical statements, as done by \citet{majumder2021rationaleinspired}. The resulting task can be described as \enquote{given a pair of sentences, identify which one does \textbf{not} make sense, and explain why}, which we refer to simply as ComVE. The resulting ComVE dataset consists of 10,000 training, 1,000 validation, and 1,000 test instances. Each instance consists of a pair of statements, a label, and three human-generated NLEs. We use all three NLEs per example only in the full test set. For training, we use up to one NLE per example, assuming a strict few-shot regime where each one NLE annotation is expensive to get. For human evaluation, we randomly sample the test dataset down to 100 instances, to save human-annotation costs.

\section{Training Details}
\label{hyperparams}

The training objective is given by cross-entropy loss with targets as described in Table~\ref{tab:input_formats}. We use the AdamW optimizer~\citep{loshchilov2018decoupled} and linear learning rate scheduler with warm-up over 10\% of the training. For all models, we fix the batch size to 16 and do a grid search over the learning rate values and the number of training epochs. For all WinoGrande models, we search over the learning rate values of 3e-4, 1e-4, and 3e-5, whereas for ComVE we search over 1e-3, 3e-4, 1e-4, and 3e-5. For e-SNLI, we train on 1, 2, 3, and 5 epochs. For WinoGrande, we train on 1, 2, 3, 5, 7, 9, and 11 epochs, and for ComVE, we train on 1, 2, 3, 5, 7, 10, and 13 epochs. When few-shot fine-tuning with NLEs, we train on 1, 2, 3, 5, 7, 10, 13, 17, 21, and 26 epochs. Multi-task learning always uses the hyperparameter range of the larger dataset. No early stopping is needed, because we use a learning rate scheduler and the number of training epochs is a hyperparameter. We do not use gradual unfreezing~\cite{howard-ruder-2018-universal_2}, because it has been shown that it does not help when applied to the T5 language model~\cite{t5}.

\begin{table*}[ht]
\centering
\scalebox{0.80}{
\begin{tabular}{l
>{\centering\arraybackslash}m{1.5cm} >{\centering\arraybackslash}m{1.5cm}
>{\centering\arraybackslash}m{3.5cm}
>{\centering\arraybackslash}m{1.5cm}
}
\hline
\textbf{Models} & \textbf{Num epochs} & \textbf{Learning rate} & \textbf{Criterion} & \textbf{Best value} \\
\hline
e-SNLI       & 3 & 3e-4 & e-SNLI dev NLE ppl & 2.192 \\
(e-SNLI, SNLI) & 3 & 3e-4 & e-SNLI dev NLE ppl & 2.199 \\

\hline
\textbf{WinoGrande Models} & & & & \\
\hline
(e-SNLI, WinoGrande)   & 5 & 1e-4 & WG-dev acc &  83.2\% \\
e-SNLI--WinoGrande  & 7 & 3e-4 & WG-dev acc & 81.0\% \\
WinoGrande          & 5 & 1e-4 & WG-dev acc & 85.1\% \\
\hline
CD--fine-tune  & 21 & 3e-4 & WG-dev NLE ppl & 4.665 \\
CD--union         & 5 & 1e-4 & WG-dev NLE ppl & 4.945 \\
WT5--fine-tune & 11 & 3e-4 & WG-dev acc & 80.8\% \\
WT5 & 5 & 1e-4 & WG-dev acc & 83.4\% \\
\hline
M1  & 3 & 3e-5 & WG-dev NLE ppl & 4.815 \\
M2  & 5 & 1e-4 & WG-dev NLE ppl & 5.419 \\
M3  & 10 & 3e-4 & WG-dev NLE ppl & 4.401 \\
M4 & 17 & 3e-4 & WG-dev NLE ppl & 5.022 \\
\hline

\textbf{ComVE Models} & & & & \\
\hline
(e-SNLI, ComVE) & 3 & 3e-4 & ComVE dev acc & 82.8\% \\
e-SNLI--ComVE & 7 & 3e-4 & ComVE dev acc & 86.8\% \\
ComVE & 5 & 3e-4 & ComVE dev acc & 88.4\% \\
\hline
CD--fine-tune & 13 & 3e-4 & ComVE dev NLE ppl & 5.170 \\
CD--union & 5 & 1e-4 & ComVE dev NLE ppl* & 9.294 \\
WT5--fine-tune & 10 & 3e-4 & ComVE dev acc & 87.0\% \\
WT5 & 5 & 1e-4 & ComVE dev acc & 84.4\% \\
\hline
M1 & 5 & 1e-4 & ComVE dev NLE ppl & 7.886 \\
M2 & 1 & 1e-3 & ComVE dev NLE ppl & 7.970 \\
M3 & 5 & 1e-3 & ComVE dev NLE ppl & 4.958 \\
M4 & 5 & 1e-3 & ComVE dev NLE ppl & 5.002 \\

\hline
\end{tabular}}
\caption{\label{tab:best_hyp}
Best hyperparameters for all trained models (including the intermediary models), along with the corresponding criterion used for model selection, and the best dev result value w.r.t.\ that criterion. The datasets in brackets denotes the model obtained by fine-tuning T5 on the union of those datasets; dataset1--dataset2 denotes subsequent fine-tuning on dataset1, then on dataset2. *--subject to the dev accuracy being large enough~($>75$\%).
}
\end{table*}

At each stage of training, the best hyperparameter combinations are selected via grid search by either the perplexity relative to target NLEs on the dev set of the child task, by dev accuracy on the child dataset, or by NLE perplexity on the e-SNLI dev set, whichever is most suitable. The selection criteria for each model, along with the best hyperparameters are given in Table~\ref{tab:best_hyp}. Note that the WG-dev accuracy in Table~\ref{tab:best_hyp} is much higher than the corresponding WG-test accuracy in Table~\ref{tab:wg_comve_models}, because WG-dev is sampled from the training dataset of WinoGrande, whereas WG-test is the original WinoGrande development set, which is filtered to increase its difficulty \cite{winogrande}. Model-generated explanations are obtained via beam search with a beam width of 5.

\section{Human Evaluation}
\label{appendix:human_evaluation}

As suggested by \citet{kayser2021evil}, for each example, the annotators are provided with two (shuffled) NLEs, one from a model and one ground-truth from the test set. This serves for mentally grounding the annotator's score of the model-generated NLE. 

Additionally, there are multiple checks placed in the data collection form to ensure high-quality annotations. Most notably, in each group of 10 instances, at least 90\% of the labels have to be answered correctly, and at least 90\% of the ground-truth NLEs have to be annotated by Yes or Weak~Yes. The final check requires that at most 80\% of the model-generated NLEs should be annotated by Yes or Weak~Yes. We included this check to ensure that the annotators are more critical, and we estimated this threshold manually. These are reasonable assumptions for both WinoGrande and ComVE, judging by the quality of the ground-truth and model-generated NLEs. 

We had 130 annotators for ComVE and 113 for WinoGrande. Most of the annotators annotated only ten model-generated NLEs each. To further ensure high-quality annotations, we re-annotated all the instances of the annotators who annotated many instances (more than $60$ for WinoGrande and more than $100$ for ComVE) but selected more than five wrong shortcomings from a sample of ten random instances, after manual inspection. We found two such annotators for ComVE and one for WinoGrande. The annotators were paid 1\$ per 10 pairs of NLEs.

\begin{table*}
\centering
\scalebox{0.8}{
\begin{tabular}{m{2.4cm}
>{\centering\arraybackslash}m{0.9cm}
>{\centering\arraybackslash}m{0.9cm}
>{\centering\arraybackslash}m{1.0cm}
>{\centering\arraybackslash}m{1.0cm}
>{\centering\arraybackslash}m{0.8cm}
>{\centering\arraybackslash}m{1.5cm}
>{\centering\arraybackslash}m{1.8cm}
>{\centering\arraybackslash}m{1.5cm}
>{\centering\arraybackslash}m{1.5cm}
>{\centering\arraybackslash}m{1.2cm}
}
\hline
    \textbf{WinoGrande Model} & \textbf{NLE score} & \textbf{Yes\%} & \textbf{Weak Yes\%} & \textbf{Weak No\%} & \textbf{No\%} & \textbf{Does not make sense\%} & \textbf{Insufficient justification\%} & \textbf{Irrelevant to the schema\%} & \textbf{Too trivial\%} & \textbf{None\%} \\
    \hline
    CD--fine-tune & 34.7 & 17.5 & 20.1 & 11.6 & 50.8 & 32.0 & 37.0 & 4.0 & 7.5 & 19.5 \\
    CD--union & 35.9 & 20.7 & 15.2 & 15.2 & 49.0 & 33.8 & 32.4 & 5.5 & 6.4 & 21.9 \\
    WT5--fine-tune & 8.7 & 4.6 & 4.1 & 4.1 & 87.2 & 60.8 & 20.3 & 10.6 & 4.1 & 4.1 \\
    WT5 & 8.3 & 4.8 & 3.0 & 4.2 & 87.9 & 71.1 & \textbf{12.8} & 9.6 & \textbf{2.1} & 4.3 \\
    \hline
    M1 & 28.3 & 14.3 & 14.3 & 13.6 & 57.8 & \textbf{28.0} & 39.5 & 8.9 & 4.5 & 19.1 \\
    M2 & \textbf{44.1} & \textbf{25.9} & 18.0 & 18.5 & \textbf{37.6} & \textbf{28.1} & 33.2 & 6.5 & 4.0 & \textbf{28.1} \\
    M3 & 29.6 & 15.4 & 14.8 & 13.0 & 56.8 & 43.7 & 29.3 & 6.9 & 2.3 & 17.8 \\
    M4 & 41.9 & 22.6 & 22.6 & 12.8 & 42.1 & 34.3 & 33.3 & \textbf{2.5} & 6.9 & 23.0 \\
    
    \hline
    \textbf{ComVE Model} & \textbf{NLE score} & \textbf{Yes\%} & \textbf{Weak Yes\%} & \textbf{Weak No\%} & \textbf{No\%} & \textbf{Does not make sense\%} & \textbf{Insufficient justification\%} & \textbf{Irrelevant to the schema\%} & \textbf{Too trivial\%} & \textbf{None\%} \\
    \hline
    CD--fine-tune & 31.4 & 25.4 & 7.2 & 3.8 & 63.6 & 26.9 & 32.3 & 12.5 & 3.6 & 24.7 \\
    CD--union & 27.7 & 23.6 & 4.2 & 3.8 & 68.4 & 39.8 & 24.6 & 10.2 & \textbf{2.7} & 22.7 \\
    WT5--fine-tune & 28.9 & 20.0 & 11.8 & 3.1 & 65.1 & 30.7 & 37.9 & \textbf{8.9} & 3.6 & 18.9 \\
    WT5 & 23.9 & 15.3 & 10.2 & 5.6 & 69.0 & 36.9 & 31.7 & 11.9 & 5.2 & 14.3 \\
    \hline
    M1 & 40.2 & 28.5 & 14.6 & 6.1 & 50.8 & 22.1 & 29.0 & 18.1 & 4.7 & 26.1 \\
    M2 & 40.6 & 27.4 & 17.7 & 4.2 & 50.6 & 23.9 & 33.5 & 10.4 & 4.4 & 27.9 \\
    M3 & 38.6 & 30.3 & 8.8 & 7.5 & 53.5 & 32.5 & \textbf{21.7} & 12.0 & 4.4 & 29.3 \\
    M4 & \textbf{48.5} & \textbf{36.7} & 14.3 & 6.8 & \textbf{42.2} & \textbf{18.8} & 28.2 & 13.1 & 2.9 & \textbf{37.1} \\
\hline
\end{tabular}}
\caption{\label{tab:wg_comve_human_evaluation}
Human annotations of the correctly-classified NLEs generated by models with WinoGrande and ComVE as child tasks (CT). The columns Yes, Weak Yes, Weak No, and No present the percentages of NLE validity scores given by the human annotators. The last five columns present the shortcomings provided by the human annotators. Best results are in bold. We do not bold the Weak Yes and Weak No, since it is not clear that higher/lower is better. 
}
\end{table*}

\begin{figure*}
  \includegraphics[width=425pt]{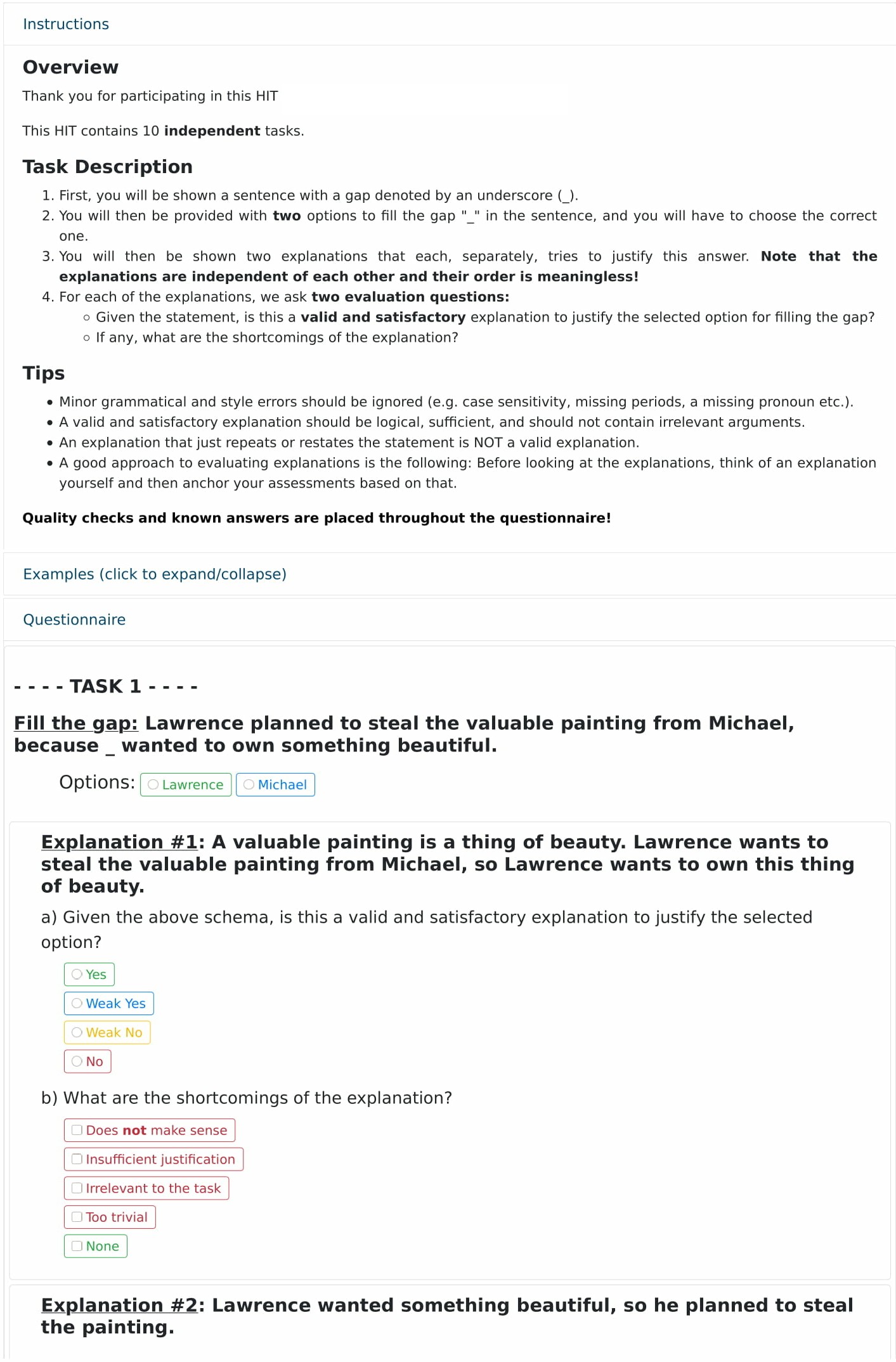}
  \caption{ \label{fig:wg_template}
  WinoGrande data collection template. There are two explanations per task.}
\end{figure*}

Below are full-page screenshots of the data collection forms that we used for WinoGrande (Figure~\ref{fig:wg_template}) and ComVE (Figure~\ref{fig:comve_template}).

\section{Additional Results}
\label{additional_results_appendix}

Table~\ref{tab:wg_comve_human_evaluation} presents the full human evaluation results table for all models which includes the separate Yes, Weak~Yes, Weak~No, and No scores. 
Table~\ref{tab:wg_comve_human_evaluation} also summarizes, for each model, the shortcomings that the human annotators found in the model-generated NLEs. The annotated shortcomings of the NLEs are informative of the issues that current generated NLEs have.

\section{Examples of Model-Generated NLEs}
\label{all_nle_examples}

\begin{figure*}
  \includegraphics[width=425pt]{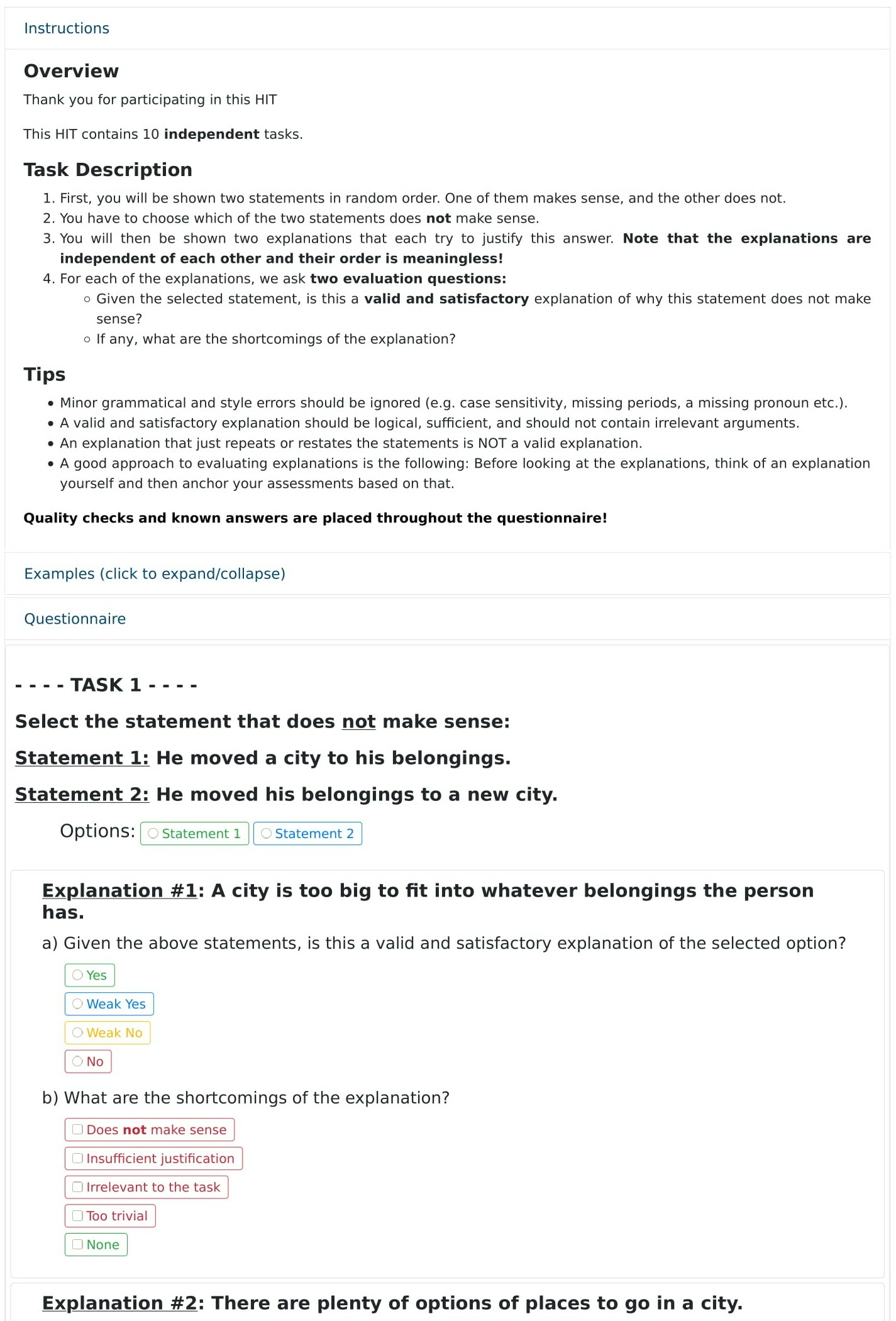}
  \caption{ \label{fig:comve_template}
  ComVE data collection template. There are two explanations per task.}
\end{figure*}

In the twelve tables below Figure~\ref{fig:comve_template} are the answers and NLEs for each child task (WinoGrande and ComVE) and for all eight compared models on the first six examples (out of the 100 that were evaluated). The first six tables present six examples for WinoGrande, whereas the second six tables are for ComVE.

\clearpage

\begin{table*}
\centering
\scalebox{0.86}{\begin{tabular}{ | >{\raggedright\arraybackslash}m{3cm} | >{\arraybackslash}m{12cm}|}
\hline
\multicolumn{2}{|>{\raggedright}m{15cm}|}{\textbf{WinoGrande schema:} The \underline{\hspace{0.5cm}} car was a pleasure to drive but the red car was not as the blue car was very expensive. \textbf{Options:} blue, red. \textbf{Ground-truth:} blue. } \\ 
\hline
\textbf{Model} & \textbf{Output} \\ 
\hline
CD--fine-tune & \textbf{Prediction:} red. \textbf{NLE:} The blue car is expensive and the blue car is more expensive. \\ \hline
CD--union & \textbf{Prediction:} red. \textbf{NLE:} The blue car is more expensive than the red car, hence it is more expensive to drive. \\ \hline
WT5--fine-tune & \textbf{Prediction:} red. \textbf{NLE:} The car is described in more detail than in the second sentence. Blue \\ \hline
WT5 & \textbf{Prediction:} blue. \textbf{NLE:} Red is a different color than blue. \\ \hline
M1 & \textbf{Prediction:} red. \textbf{NLE:} The car was a pleasure to drive but the blue car was not as the blue car was very expensive. \\ \hline
M2 & \textbf{Prediction:} red. \textbf{NLE:} The red car is more expensive than the blue car. \\ \hline
M3 & \textbf{Prediction:} blue. \textbf{NLE:} The blue car is a pleasure to drive and the red car is very expensive. \\ \hline
M4 & \textbf{Prediction:} blue. \textbf{NLE:} The car is a pleasure to drive but the red car is red. \\ \hline
\end{tabular}} 
\end{table*}

\begin{table*}
\centering
\scalebox{0.86}{\begin{tabular}{ | >{\raggedright\arraybackslash}m{3cm} | >{\arraybackslash}m{12cm}|}
\hline
\multicolumn{2}{|>{\raggedright}m{15cm}|}{\textbf{WinoGrande schema:} Logan ended the one-side relationship they were having with Robert, which made \underline{\hspace{0.5cm}} feel relieved. \textbf{Options:} Logan, Robert. \textbf{Ground-truth:} Logan. } \\ 
\hline
\textbf{Model} & \textbf{Output} \\ 
\hline
CD--fine-tune & \textbf{Prediction:} Robert. \textbf{NLE:} Logan ended the one-side relationship they were having with Robert, which made Logan feel relieved. \\ \hline
CD--union & \textbf{Prediction:} Logan. \textbf{NLE:} Robert has a one-sided relationship with Logan, which makes him feel relieved. \\ \hline
WT5--fine-tune & \textbf{Prediction:} Logan. \textbf{NLE:} The Logan \\ \hline
WT5 & \textbf{Prediction:} Logan. \textbf{NLE:} Robert \\ \hline
M1 & \textbf{Prediction:} Robert. \textbf{NLE:} Logan ended the one-side relationship they were having with Robert, which made Logan feel relieved \\ \hline
M2 & \textbf{Prediction:} Logan. \textbf{NLE:} The fact that Logan ended the one-sided relationship with Robert makes him feel relieved. \\ \hline
M3 & \textbf{Prediction:} Robert. \textbf{NLE:} Logan's feelings are relieved because he ended the relationship with Robert. \\ \hline
M4 & \textbf{Prediction:} Logan. \textbf{NLE:} Logan ending a one-side relationship makes Logan feel relieved. \\ \hline
\end{tabular}} 
\end{table*}

\begin{table*}
\centering
\scalebox{0.86}{\begin{tabular}{ | >{\raggedright\arraybackslash}m{3cm} | >{\arraybackslash}m{12cm}|}
\hline
\multicolumn{2}{|>{\raggedright}m{15cm}|}{\textbf{WinoGrande schema:} Fiction books were interesting and easy to read for  Logan but not Brett because \underline{\hspace{0.5cm}} enjoyed real tales. \textbf{Options:} Logan, Brett. \textbf{Ground-truth:} Brett. } \\ 
\hline
\textbf{Model} & \textbf{Output} \\ 
\hline
CD--fine-tune & \textbf{Prediction:} Brett. \textbf{NLE:} Fiction books were interesting and easy to read for Logan, but Brett enjoyed real tales more than fiction books. \\ \hline
CD--union & \textbf{Prediction:} Brett. \textbf{NLE:} Fiction books were interesting and easy to read for Logan, while real tales were more difficult to read for Brett. \\ \hline
WT5--fine-tune & \textbf{Prediction:} Logan. \textbf{NLE:} There is nothing to indicate that he is enjoying real stories or that he has been referred to as 'Boy' \\ \hline
WT5 & \textbf{Prediction:} Logan. \textbf{NLE:} The Logan may not be Logan. Brett may not be Brett. Logan may not be Logan. Brett may not be Brett. Logan may not be Logan. Brett may not be Brett. Logan may not be Logan. Brett may not be Brett. Loga [\ldots] \\ \hline
M1 & \textbf{Prediction:} Brett. \textbf{NLE:} Fiction books were interesting and easy to read for Logan but not Brett because Brett enjoyed real tales \\ \hline
M2 & \textbf{Prediction:} Brett. \textbf{NLE:} Fiction books are not real tales. \\ \hline
M3 & \textbf{Prediction:} Logan. \textbf{NLE:} Fiction books are interesting and easy to read for Logan, so Logan is more likely to enjoy real tales than Brett. \\ \hline
M4 & \textbf{Prediction:} Logan. \textbf{NLE:} Fiction books are not real tales. \\ \hline
\end{tabular}} 
\end{table*}

\begin{table*}
\centering
\scalebox{0.86}{\begin{tabular}{ | >{\raggedright\arraybackslash}m{3cm} | >{\arraybackslash}m{12cm}|}
\hline
\multicolumn{2}{|>{\raggedright}m{15cm}|}{\textbf{WinoGrande schema:} Erin ordered a martini when Tanya only ordered a tonic water with lemon, because \underline{\hspace{0.5cm}} was pregnant. \textbf{Options:} Erin, Tanya. \textbf{Ground-truth:} Tanya. } \\ 
\hline
\textbf{Model} & \textbf{Output} \\ 
\hline
CD--fine-tune & \textbf{Prediction:} Tanya. \textbf{NLE:} Erin ordered a martini when Tanya ordered tonic water with lemon because she was pregnant. \\ \hline
CD--union & \textbf{Prediction:} Erin. \textbf{NLE:} Erin ordered a martini when Tanya ordered a tonic water with lemon, because Erin was pregnant. \\ \hline
WT5--fine-tune & \textbf{Prediction:} Erin. \textbf{NLE:} The Tanya is pregnant. Tanya is pregnant. Tanya \\ \hline
WT5 & \textbf{Prediction:} Tanya. \textbf{NLE:} Erin and Tanya are two different people. Erin and Tanya are two different people. Erin and Tanya are two different people. Erin and Tanya are two different people. Erin and Tanya are two different peo [\ldots] \\ \hline
M1 & \textbf{Prediction:} Tanya. \textbf{NLE:} Erin ordered a martini when Tanya only ordered a tonic water with lemon, because \underline{\hspace{0.3cm}} was pregnant. \\ \hline
M2 & \textbf{Prediction:} Tanya. \textbf{NLE:} Erin ordering a martini implies Erin ordering a martini, and Tanya ordering a tonic implies Erin is pregnant. \\ \hline
M3 & \textbf{Prediction:} Tanya. \textbf{NLE:} Erin ordered a martini \\ \hline
M4 & \textbf{Prediction:} Erin. \textbf{NLE:} Erin ordering a martini implies Erin is pregnant. Erin ordering a tonic water implies Erin is pregnant. Erin ordering a martini implies Erin ordering a martini. Erin ordering a tonic water implies Eri [\ldots] \\ \hline
\end{tabular}} 
\end{table*}

\begin{table*}
\centering
\scalebox{0.86}{\begin{tabular}{ | >{\raggedright\arraybackslash}m{3cm} | >{\arraybackslash}m{12cm}|}
\hline
\multicolumn{2}{|>{\raggedright}m{15cm}|}{\textbf{WinoGrande schema:} The game on television was very important for Ian but not Dennis because \underline{\hspace{0.5cm}} was into music instead. \textbf{Options:} Ian, Dennis. \textbf{Ground-truth:} Dennis. } \\ 
\hline
\textbf{Model} & \textbf{Output} \\ 
\hline
CD--fine-tune & \textbf{Prediction:} Dennis. \textbf{NLE:} The game on television was very important for Ian but not for Dennis because he was into music instead. \\ \hline
CD--union & \textbf{Prediction:} Dennis. \textbf{NLE:} The game on television was very important for Ian, but not for Dennis, because he was into music instead. \\ \hline
WT5--fine-tune & \textbf{Prediction:} Dennis. \textbf{NLE:} He is playing music instead of playing music instead of playing music instead of playing music instead of playing music instead of playing music instead of playing music instead of playing music inste [\ldots] \\ \hline
WT5 & \textbf{Prediction:} Dennis. \textbf{NLE:} The Ian can't be Dennis. Dennis can't be Ian. Dennis can't be Ian. Dennis can't be Dennis. Dennis can't be Ian. Dennis can't be Ian. Dennis can't be Dennis. Dennis can't be Ian. Dennis can't be Dennis [\ldots] \\ \hline
M1 & \textbf{Prediction:} Dennis. \textbf{NLE:} The game on television was very important for Ian but not Dennis because he was into music instead. \\ \hline
M2 & \textbf{Prediction:} Dennis. \textbf{NLE:} The game on television is more important for Ian than Dennis because Dennis is more likely to be into music instead. \\ \hline
M3 & \textbf{Prediction:} Dennis. \textbf{NLE:} The game on television is very important for Ian \\ \hline
M4 & \textbf{Prediction:} Dennis. \textbf{NLE:} The game on television is very important for Ian, so he is not into music. \\ \hline
\end{tabular}} 
\end{table*}

\begin{table*}
\centering
\scalebox{0.86}{\begin{tabular}{ | >{\raggedright\arraybackslash}m{3cm} | >{\arraybackslash}m{12cm}|}
\hline
\multicolumn{2}{|>{\raggedright}m{15cm}|}{\textbf{WinoGrande schema:} It was easier for the diner to follow their budget than the food truck because the \underline{\hspace{0.5cm}} had more money to spend. \textbf{Options:} diner, food truck. \textbf{Ground-truth:} diner. } \\ 
\hline
\textbf{Model} & \textbf{Output} \\ 
\hline
CD--fine-tune & \textbf{Prediction:} diner. \textbf{NLE:} The diner has more money to spend than the food truck. \\ \hline
CD--union & \textbf{Prediction:} diner. \textbf{NLE:} The diner has more money to spend than the food truck, so it is easier for them to follow their budget. \\ \hline
WT5--fine-tune & \textbf{Prediction:} diner. \textbf{NLE:} food truck \\ \hline
WT5 & \textbf{Prediction:} food truck. \textbf{NLE:} diner \\ \hline
M1 & \textbf{Prediction:} diner. \textbf{NLE:} The diner had more money to spend than the food truck. \\ \hline
M2 & \textbf{Prediction:} diner. \textbf{NLE:} The diner has more money to spend than the food truck. \\ \hline
M3 & \textbf{Prediction:} diner. \textbf{NLE:} The diner has more money to spend than the food truck. \\ \hline
M4 & \textbf{Prediction:} diner. \textbf{NLE:} More money to spend is a rephrasing of more money to spend. \\ \hline
\end{tabular}} 
\end{table*}

\begin{table*}
\centering
\scalebox{0.86}{\begin{tabular}{ | >{\raggedright\arraybackslash}m{3cm} | >{\arraybackslash}m{12cm}|}
\hline
\multicolumn{2}{|>{\raggedright}m{15cm}|}{\textbf{ComVE Statement 1:} He inserts his thumb into the car ignition switch. \textbf{Statement 2:} He inserts the keys into the car ignition switch. \textbf{Nonsensical statement:} Statement 1. } \\ 
\hline
\textbf{Model} & \textbf{Output} \\ 
\hline
CD--fine-tune & \textbf{Prediction:} Statement 1. \textbf{NLE:} A thumb is not inserted into the car ignition switch. \\ \hline
CD--union & \textbf{Prediction:} Statement 1. \textbf{NLE:} It's not a car ignition switch, it's a car ignition switch, it's a car ignition switch, it's a car ignition switch, it's a car ignition switch, it's a car ignition switch, it's a car ignition switch,  [\ldots] \\ \hline
WT5--fine-tune & \textbf{Prediction:} Statement 1. \textbf{NLE:} SIDEWALK INSTALLS THE KEYS INTO THE CAR INFUSION SHUT. \\ \hline
WT5 & \textbf{Prediction:} Statement 1. \textbf{NLE:} the keys are not the same as the thumb \\ \hline
M1 & \textbf{Prediction:} Statement 1. \textbf{NLE:} One cannot insert his thumb into the car ignition switch. \\ \hline
M2 & \textbf{Prediction:} Statement 1. \textbf{NLE:} One cannot insert his thumb into the car ignition switch. \\ \hline
M3 & \textbf{Prediction:} Statement 1. \textbf{NLE:} You can't use your thumb to insert keys into the car ignition switch. \\ \hline
M4 & \textbf{Prediction:} Statement 1. \textbf{NLE:} A thumb is not used to insert keys into a car ignition switch \\ \hline
\end{tabular}} 
\end{table*}

\begin{table*}
\centering
\scalebox{0.86}{\begin{tabular}{ | >{\raggedright\arraybackslash}m{3cm} | >{\arraybackslash}m{12cm}|}
\hline
\multicolumn{2}{|>{\raggedright}m{15cm}|}{\textbf{ComVE Statement 1:} She was met with an accident and was rushed to the hospital \textbf{Statement 2:} She was met with an accident and was rushed to the library \textbf{Nonsensical statement:} Statement 2. } \\ 
\hline
\textbf{Model} & \textbf{Output} \\ 
\hline
CD--fine-tune & \textbf{Prediction:} Statement 2. \textbf{NLE:} The library is closed to the public \\ \hline
CD--union & \textbf{Prediction:} Statement 2. \textbf{NLE:} She was met with an accident and was rushed to the library \\ \hline
WT5--fine-tune & \textbf{Prediction:} Statement 2. \textbf{NLE:} She was met with an accident and was rushed to the hospital 1 explanation: \\ \hline
WT5 & \textbf{Prediction:} Statement 2. \textbf{NLE:} . \\ \hline
M1 & \textbf{Prediction:} Statement 2. \textbf{NLE:} The library is not the only place where a person can have an accident. \\ \hline
M2 & \textbf{Prediction:} Statement 2. \textbf{NLE:} A library is not a hospital. \\ \hline
M3 & \textbf{Prediction:} Statement 2. \textbf{NLE:} A library is not a hospital. \\ \hline
M4 & \textbf{Prediction:} Statement 2. \textbf{NLE:} A library is not a hospital. \\ \hline
\end{tabular}} 
\end{table*}

\begin{table*}
\centering
\scalebox{0.86}{\begin{tabular}{ | >{\raggedright\arraybackslash}m{3cm} | >{\arraybackslash}m{12cm}|}
\hline
\multicolumn{2}{|>{\raggedright}m{15cm}|}{\textbf{ComVE Statement 1:} She came off sailing across the pacific ocean \textbf{Statement 2:} She came off walking across the pacific ocean \textbf{Nonsensical statement:} Statement 2. } \\ 
\hline
\textbf{Model} & \textbf{Output} \\ 
\hline
CD--fine-tune & \textbf{Prediction:} Statement 2. \textbf{NLE:} She came off sailing across the pacific ocean \\ \hline
CD--union & \textbf{Prediction:} Statement 2. \textbf{NLE:} She came off sailing across the pacific ocean Sailing across the pacific ocean Sailing across the pacific ocean Sailing across the pacific ocean Sailing across the pacific ocean Sailing across the pac [\ldots] \\ \hline
WT5--fine-tune & \textbf{Prediction:} Statement 2. \textbf{NLE:} She came off sailing across the pacific ocean Sentence 1: \\ \hline
WT5 & \textbf{Prediction:} Statement 2. \textbf{NLE:} . \\ \hline
M1 & \textbf{Prediction:} Statement 2. \textbf{NLE:} A person can't be sailing and walking at the same time. \\ \hline
M2 & \textbf{Prediction:} Statement 2. \textbf{NLE:} sailing is not the same as walking \\ \hline
M3 & \textbf{Prediction:} Statement 2. \textbf{NLE:} Sailing and walking are two different things. \\ \hline
M4 & \textbf{Prediction:} Statement 2. \textbf{NLE:} sailing is not the same as walking \\ \hline
\end{tabular}} 
\end{table*}

\begin{table*}
\centering
\scalebox{0.86}{\begin{tabular}{ | >{\raggedright\arraybackslash}m{3cm} | >{\arraybackslash}m{12cm}|}
\hline
\multicolumn{2}{|>{\raggedright}m{15cm}|}{\textbf{ComVE Statement 1:} Leena daily eats two bananas which are good for health. \textbf{Statement 2:} Leena eats almost 1000 bananas daily as his morning breakfast. \textbf{Nonsensical statement:} Statement 2. } \\ 
\hline
\textbf{Model} & \textbf{Output} \\ 
\hline
CD--fine-tune & \textbf{Prediction:} Statement 2. \textbf{NLE:} Leena eats two bananas a day which are good for health. \\ \hline
CD--union & \textbf{Prediction:} Statement 2. \textbf{NLE:} Bananas are good for health and you can't eat them in the morning. \\ \hline
WT5--fine-tune & \textbf{Prediction:} Statement 2. \textbf{NLE:} Leena daily eats two bananas which are good for health 1 explanation: \\ \hline
WT5 & \textbf{Prediction:} Statement 2. \textbf{NLE:} . \\ \hline
M1 & \textbf{Prediction:} Statement 2. \textbf{NLE:} Leena daily eats two bananas which are good for health. is same as Leena eats almost 1000 bananas daily as his morning breakfast \\ \hline
M2 & \textbf{Prediction:} Statement 2. \textbf{NLE:} The bananas are good for health. \\ \hline
M3 & \textbf{Prediction:} Statement 2. \textbf{NLE:} The bananas are not good for health, they are good for health. \\ \hline
M4 & \textbf{Prediction:} Statement 2. \textbf{NLE:} The bananas are good for health. \\ \hline
\end{tabular}} 
\end{table*}

\begin{table*}
\centering
\scalebox{0.86}{\begin{tabular}{ | >{\raggedright\arraybackslash}m{3cm} | >{\arraybackslash}m{12cm}|}
\hline
\multicolumn{2}{|>{\raggedright}m{15cm}|}{\textbf{ComVE Statement 1:} The fire will burn you if you touch it. \textbf{Statement 2:} The fire will cool you if you touch it. \textbf{Nonsensical statement:} Statement 2. } \\ 
\hline
\textbf{Model} & \textbf{Output} \\ 
\hline
CD--fine-tune & \textbf{Prediction:} Statement 2. \textbf{NLE:} The fire will burn you if you touch it. \\ \hline
CD--union & \textbf{Prediction:} Statement 2. \textbf{NLE:} The fire will burn you if you touch it and will cool you if you touch it \\ \hline
WT5--fine-tune & \textbf{Prediction:} Statement 2. \textbf{NLE:} The fire will burn you if you touch it. \\ \hline
WT5 & \textbf{Prediction:} Statement 2. \textbf{NLE:} the fire will burn you if you touch it. \\ \hline
M1 & \textbf{Prediction:} Statement 2. \textbf{NLE:} The fire will burn you is not the same as the fire that will cool you. \\ \hline
M2 & \textbf{Prediction:} Statement 2. \textbf{NLE:} The fire will burn you if you touch it \\ \hline
M3 & \textbf{Prediction:} Statement 1. \textbf{NLE:} Fire can't burn you \\ \hline
M4 & \textbf{Prediction:} Statement 2. \textbf{NLE:} The fire will burn you. \\ \hline
\end{tabular}} 
\end{table*}


\begin{table*}\label{tab-last}
\centering
\scalebox{0.86}{\begin{tabular}{ | >{\raggedright\arraybackslash}m{3cm} | >{\arraybackslash}m{12cm}|}
\hline
\multicolumn{2}{|>{\raggedright}m{15cm}|}{\textbf{ComVE Statement 1:} Salsa is for hair \textbf{Statement 2:} Brushes are for hair \textbf{Nonsensical statement:} Statement 1. } \\ 
\hline
\textbf{Model} & \textbf{Output} \\ 
\hline
CD--fine-tune & \textbf{Prediction:} Statement 1. \textbf{NLE:} Salsa is for hair \\ \hline
CD--union & \textbf{Prediction:} Statement 1. \textbf{NLE:} Salsa is for hair \\ \hline
WT5--fine-tune & \textbf{Prediction:} Statement 1. \textbf{NLE:} SENTENCE 1: \\ \hline
WT5 & \textbf{Prediction:} Statement 1. \textbf{NLE:} : \\ \hline
M1 & \textbf{Prediction:} Statement 1. \textbf{NLE:} Brushes are for hair is not the same as a Salsa \\ \hline
M2 & \textbf{Prediction:} Statement 1. \textbf{NLE:} Salsa is for hair \\ \hline
M3 & \textbf{Prediction:} Statement 1. \textbf{NLE:} Brushes are for hair is a rephrasing of brushes are for hair. \\ \hline
M4 & \textbf{Prediction:} Statement 1. \textbf{NLE:} Salsa is for hair \\ \hline
\end{tabular}} 
\end{table*}